\documentclass[10pt,twocolumn,letterpaper]{article}

\usepackage{cvpr}              %

\usepackage{graphicx}
\usepackage{amsmath}
\usepackage{amssymb}
\usepackage{booktabs}

\usepackage{url}
\usepackage{soul}

\usepackage{caption}
\usepackage{graphicx} %
\usepackage[hyphens]{url}  %
\usepackage{enumitem}

\usepackage{xspace}
\usepackage{multirow}
\usepackage{booktabs}
\usepackage{xcolor}
\usepackage[pagebackref,breaklinks,colorlinks=true,citecolor=blue]{hyperref}
\usepackage[ruled,vlined]{algorithm2e}

\usepackage{amsmath,amsfonts,bm}

\def\eqref#1{equation~\ref{#1}}

\def\1{\bm{1}}

\def\vk{{\bm{k}}}

\def\vx{{\bm{x}}}

\DeclareMathAlphabet{\mathsfit}{\encodingdefault}{\sfdefault}{m}{sl}
\SetMathAlphabet{\mathsfit}{bold}{\encodingdefault}{\sfdefault}{bx}{n}

\def\sR{{\mathbb{R}}}

\newcommand{\zizhaoz}[1]{\textcolor{black}{#1}}
\newcommand{\zifeng}[1]{\textcolor{black}{#1}}
\newcommand{\final}[1]{\textcolor{black}{#1}}
\newcommand{\finalx}[1]{\textcolor{black}{#1}}
\newcommand{\cam}[1]{\textcolor{black}{#1}}

\usepackage{amssymb}%
\usepackage{pifont}%

\newcommand{\method}{Learning to Prompt for Continual Learning\xspace}
\newcommand{\methodabbr}{L2P\xspace}

\def\ie{\emph{i.e.}\xspace}

\def\pp{\mathbf{P}}
\def\kk{\mathbf{K}}

\usepackage[capitalize]{cleveref}
\crefname{section}{Sec.}{Secs.}
\Crefname{section}{Section}{Sections}
\Crefname{table}{Table}{Tables}
\crefname{table}{Tab.}{Tabs.}

\newcounter{alphasect}
\def\alphainsection{0}

\let\oldsection=\section
\def\section{%
  \ifnum\alphainsection=1%
    \addtocounter{alphasect}{1}
  \fi%
\oldsection}%

\renewcommand\thesection{%
 \ifnum\alphainsection=1%
   \Alph{alphasect}%
 \else
   \arabic{section}%
 \fi%
}%

\newenvironment{alphasection}{%
  \ifnum\alphainsection=1%
    \errhelp={Let other blocks end at the beginning of the next block.}
    \errmessage{Nested Alpha section not allowed}
  \fi%
  \setcounter{alphasect}{0}
  \def\alphainsection{1}
}{%
  \setcounter{alphasect}{0}
  \def\alphainsection{0}
}%

\def\cvprPaperID{9806} %
\def\confName{CVPR}
\def\confYear{2022}

\begin{document}

\title{Learning to Prompt for Continual Learning}
\author{\textbf{Zifeng Wang\textsuperscript{1}\thanks{Work done during internship at Google Cloud AI Research.} \quad  Zizhao Zhang\textsuperscript{2} \quad  Chen-Yu Lee\textsuperscript{2} \quad  Han Zhang\textsuperscript{3} \quad Ruoxi Sun\textsuperscript{2}} \\ 
\textbf{Xiaoqi Ren\textsuperscript{2}\quad Guolong Su\textsuperscript{3}\quad Vincent Perot\textsuperscript{3} \quad Jennifer Dy\textsuperscript{1} \quad Tomas Pfister\textsuperscript{2}} \\
\textsuperscript{1}Northeastern University \quad \textsuperscript{2}Google Cloud AI
\quad \textsuperscript{3}Google Research}
\maketitle

\begin{abstract}
The mainstream paradigm behind continual learning has been to adapt the model parameters to non-stationary data distributions, where catastrophic forgetting is the central challenge. \zizhaoz{Typical methods rely on a rehearsal buffer or known task identity at test time to retrieve learned knowledge and address forgetting, while this work presents a new paradigm for continual learning that aims to train a more succinct memory system without accessing task identity at test time. Our method learns to dynamically \emph{prompt} (\methodabbr) a pre-trained model to learn tasks sequentially under different task transitions.}
\zizhaoz{In our proposed framework, prompts are small learnable parameters, which are maintained in a memory space.
The objective is to optimize prompts to instruct the model prediction and explicitly manage task-invariant and task-specific knowledge while maintaining model plasticity.}
\zifeng{We conduct comprehensive experiments under popular image classification benchmarks with different challenging continual learning settings, where \methodabbr consistently outperforms prior state-of-the-art methods. Surprisingly, \methodabbr  achieves competitive results against rehearsal-based methods even without a rehearsal buffer and is directly applicable to challenging task-agnostic continual learning. %
}
Source code is available at \url{https://github.com/google-research/l2p}. %
\end{abstract}
\vspace{-0.3cm}

\section{Introduction}
Contrary to ordinary supervised learning that trains on independent and identically distributed (i.i.d.) data, continual learning tackles the problem of training a single model on non-stationary data distributions where different classification tasks are presented sequentially.
However, since the model only has access to the current data in an individual phase of the learning cycle, it is prone to overfit on the currently available data and suffers from performance deterioration on the previously trained data due to \emph{catastrophic forgetting} \cite{mccloskey1989catastrophic}.

\begin{figure}[t]
\centering 
\includegraphics[width=0.999\linewidth]{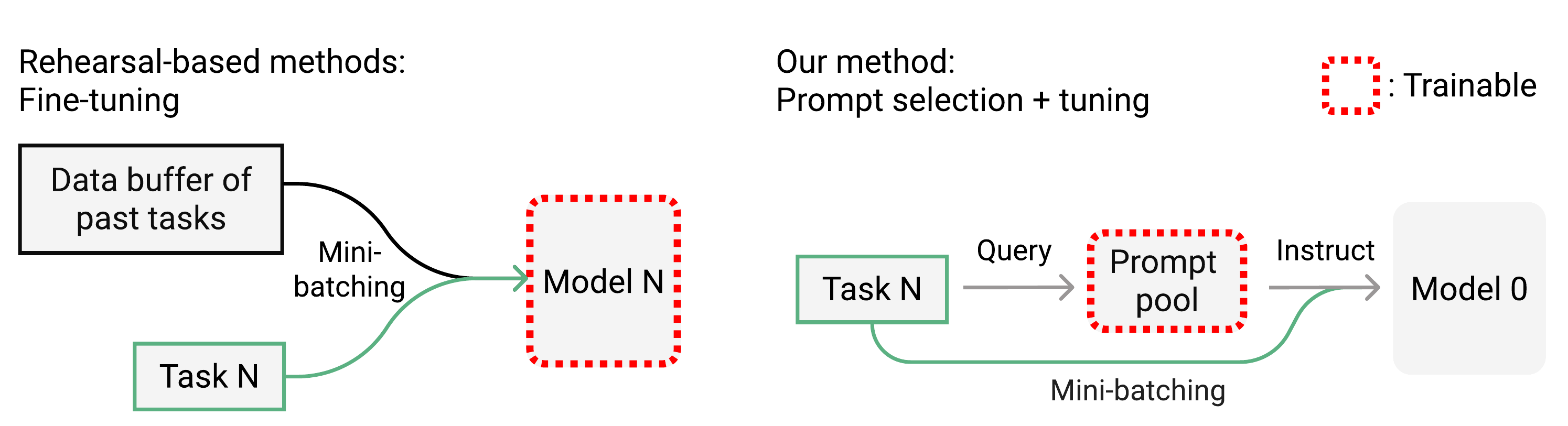} 
\vspace{-.4cm}
\caption{Overview of the \methodabbr framework. 
Compared with typical methods that adapt entire or partial model weights to tasks sequentially with a rehearsal buffer to avoid forgetting, \methodabbr uses a single backbone model and learns a prompt pool to instruct the model conditionally. \zizhaoz{Task-specific knowledge is stored inside a prompt pool, \final{thus a rehearsal buffer is no longer mandatory to mitigate forgetting}.} \zifeng{\methodabbr automatically selects and updates prompts from the pool in an instance-wise fashion, thus task identity is not required at test time.} \final{Notably, our largest prompt space is smaller than the size of one $224\times224$ image.}}
\label{fig:overview} \vspace{-.3cm}
\end{figure} 

\zifeng{
A major body of work in continual learning follows the learning paradigm by adapting the entire or partial model weights continually as the data distribution shifts, with a focus on preserving past knowledge \cite{delange2021continual ,mai2021online}. Although many types of methods attain good results, there are still critical limitations that need to be addressed. First, \final{motivated by the episodic memory in the hippocampus according to the Complementary Learning  Systems (CLS) theory~\cite{mcclelland1995there, kumaran2016learning}}, many state-of-the-art methods \cite{chaudhry2019tiny, buzzega2020dark, cha2021co2l} rely on a rehearsal buffer to re-train a portion of past examples. However, they suffer from substantial performance deterioration with smaller buffer size \cite{cha2021co2l} and 
become ineffective when a rehearsal buffer is not allowed -- for example, in real-world scenarios where data privacy matters \cite{shokri2015privacy}. 
\final{This suggests that simply buffering past data and re-train the model may not be the best approach to retrieve past knowledge.}
Without accessing a rehearsal buffer,
another branch of works \cite{li2019learn, ke2020continual, pham2020contextual} bypass the forgetting issue by assuming known task identity at test time, so that they are able to attach task-independent modules to the shared model for inference. However, knowing task identity at test time restricts practical usage. %
}

\zifeng{
The limitations of prior work bring up critical questions in continual learning \cite{farquhar2018towards, hadsell2020embracing}: (1) \final{Whether the form of episodic memory can go beyond buffering past data to more intelligent and succinct episodic memory system}? 
(2) How to automatically select relevant knowledge component for arbitrary sample without knowing its task identity?
}

\zifeng{
To answer the first question, we draw inspiration from recent advances in prompt-based learning (prompting) \cite{liu2021pre}, a new transfer learning technique in the field of natural language processing (NLP).
}
Prompting techniques design model textual inputs with templated or learnable \emph{prompt} tokens containing additional task-specific information, such that the pre-trained language model can process parameterized inputs in order to perform prompt-specific prediction \cite{lester2021power,shin2020autoprompt,li2021prefix}. 
Intuitively, prompt-based learning reformulates learning downstream tasks from directly adapting model weights to designing prompts that ``instruct'' the model to perform tasks conditionally. A prompt encodes task-specific knowledge and has the ability to utilize pre-trained frozen models more effectively than ordinary fine-tuning \cite{lester2021power,raffel2020exploring}. \zifeng{Thus, it is promising to leverage prompts to learn knowledge, \final{and further store learned knowledge,} in the continual learning context.}

\zifeng{
Nevertheless, it is not clear how to apply prompting to address the aforementioned second question  in continual learning directly: 
On one hand, if we train different prompts for different tasks in the continual learning context, test-time task identity is still required for making predictions using an appropriate task-specific prompt. On the other hand, as a transfer learning technique, the target of prompting is to make frozen pre-trained models achieve good performance on down-streaming individually, not sequentially. Therefore, if we instead maintain a single shared prompt for all tasks, the problem of catastrophic forgetting may still exist (see Section~\ref{sec:ablation}).
}

\begin{figure*}[t]
\centering 
\includegraphics[width=0.75\linewidth]{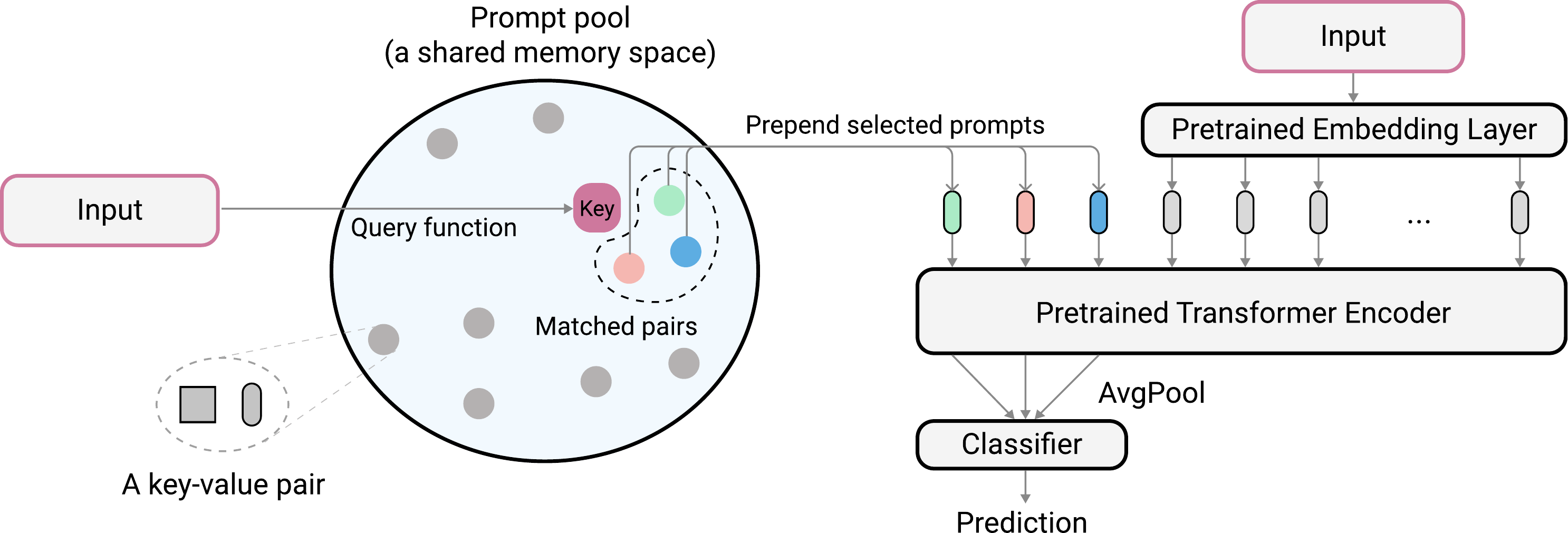} 
\vspace{-.1cm}
\caption{Illustration of \methodabbr at test time.
\zifeng{We follow the same procedure at training time:} First, \methodabbr selects a subset of prompts from a key-value paired \emph{prompt pool} based on our proposed instance-wise query mechanism. 
Then, \methodabbr prepends the selected prompts to the input tokens. Finally, \methodabbr feeds the extended tokens to the model, and optimize the prompt pool \zifeng{through the loss defined in~\eqref{eq:full_loss}}. 
The objective is learning to select and update prompts to instruct the prediction of the pre-trained backbone model. %
}
\label{fig:method}
\vspace{-.2cm}
\end{figure*}

To this end, we propose a new continual learning method called \textbf{L}earning \textbf{to} \textbf{P}rompt for Continual Learning (\textbf{\methodabbr}), \zifeng{which is orthogonal to popular rehearsal-based methods and applicable to practical continual learning scenarios without known task identity or boundaries.}
Figure~\ref{fig:overview} gives an overview of our method in contrast to typical continual learning methods.
\methodabbr leverages the representative features from pre-trained models; however, instead of tuning the parameters during the continual learning process, \methodabbr keeps the pre-trained model untouched, and instead learns a set of prompts that dynamically instruct models to solve corresponding tasks. %
Specifically, the prompts are structured in a key-value shared memory space called the prompt pool, and we design a query mechanism to dynamically lookup a subset of task-relevant prompts based on the instance-wise input features. 
The prompt pool, which is optimized jointly with the supervised loss, ensures that shared prompts encode shared knowledge for knowledge transfer, and unshared prompts encode task-specific knowledge that help maintain model plasticity. 
\zifeng{Our design explicitly decouples shared and task-specific knowledge, thus largely reducing the interference between task-specific knowledge during optimization, leading to minimal catastrophic forgetting without the necessity of a rehearsal buffer.}
The instance-wise query mechanism removes the necessity of knowing the task identity or boundaries, enabling the most challenging, yet under-investigated \emph{task-agnostic} continual learning.
The selected prompts are then prepended to the input embeddings (Figure~\ref{fig:method}), which implicitly add task-relevant instruction to pre-trained models, so that the model recalls the most relevant features to conduct corresponding tasks. 
In summary, this work makes the following contributions:
\begin{enumerate}[leftmargin=0.7cm]
    
    \item \zifeng{We propose \methodabbr, a novel continual learning framework based on prompts for continual learning,} \zizhaoz{providing a new mechanism to tackle continual learning challenges through learning a prompt pool memory space, which are served as parameterized ``instructions" for pre-trained models to learn tasks sequentially. The method is applicable to handle the most challenging task-agnostic continual learning.}
    
    \item \zifeng{We conduct comprehensive experiments to demonstrate the effectiveness of \methodabbr on multiple continual learning benchmarks, including class- and domain-incremental, and task-agnostic settings.~The proposed \methodabbr outperforms previous state-of-the-art methods consistently on all benchmarks. 
    Surprisingly, even when a rehearsal buffer is \emph{not} used,  \methodabbr still achieves competitive results against rehearsal-based methods, which is ideal in real-world scenarios when rehearsal buffer is prohibited.
    }

    \item To the best of our knowledge, we are the first to introduce the idea of prompting in the field of continual learning.~\zizhaoz{%
    We expect that our method provides a different perspective for solving frontier challenges in continual learning. %
    }

\end{enumerate}

\section{Related Work}
\zizhaoz{Here we draw connections and discuss differences between our method to related works.}

\textbf{Continual learning.} There are three main categories of recent continual learning algorithms.
\emph{Regularization-based} methods \cite{kirkpatrick2017overcoming, zenke2017continual, li2017learning, aljundi2018memory} limit the plasticity of the model by limiting the learning rate on important parameters for previous tasks. 
Although these methods address catastrophic forgetting to some extent \zifeng{without storing past examples}, they cannot get satisfactory performance under challenging settings\cite{mai2021online} or complex datasets \cite{rebuffi2017icarl, wu2019large}.

\emph{Rehearsal-based} methods \cite{chaudhry2018efficient, chaudhry2019tiny, hayes2019memory} construct a data buffer %
to save samples from older tasks to train with data from the current task. \zifeng{Based on this simple yet effective idea, many recent methods improve upon it by involving additional knowledge distillation penalties \cite{rebuffi2017icarl, wu2019large, chaudhry2020using, buzzega2020dark}, or leveraging self-supervised learning techniques \cite{cha2021co2l, pham2021dualnet}. Albeit its simplicity in concept, rehearsal-based methods achieve state-of-the-art performance on various benchmarks \cite{parisi2019continual, mai2021online}. However, the performance of rehearsal-based methods generally deteriorates with smaller buffer size \cite{cha2021co2l}, and rehearsal-based methods are eventually not applicable to scenarios where data privacy should be taken into account \cite{shokri2015privacy}. \final{Different from directly saving data from past knowledge to re-train the model, our method stores past knowledge in small learnable prompt parameters to instruct the model to deal with current task, and in turn accumulate current knowledge to the prompts.}
Our method does not need a rehearsal buffer to achieve performance close to rehearsal-based methods, and could be further improved to set a new stat of the art given a small rehearsal buffer.
}

\emph{Architecture-based} methods aim at having separate components for each task. The task-specific components can be identified by expanding the network \cite{rusu2016progressive, yoon2017lifelong, li2019learn, loo2020generalized, rao2019continual, zhao2022deep}, or attend to task-specific sub-networks \cite{mallya2018packnet, serra2018overcoming, wang2020learn, ke2020continual}. However, \cam{most methods, which require task identity to condition the network at test-time, are not applicable to more realistic class-incremental and task-agnostic settings when task identity is unknown. Some recent methods either infer task identity directly~\cite{wortsman2020supermasks}, or additionally add rehearsal buffer to bypass the problem~\cite{yan2021dynamically, pham2021dualnet}.} Nevertheless, these methods require substantial amount of additional parameters, sometimes close to the size of the full model \cite{wang2020learn, ke2020continual}. On the contrary,~\methodabbr does not require test-time task identity and only adds negligible amount of additional parameters ($\sim0.1\%$). \cam{Although \methodabbr also introduces additional prompt parameters, it has a totally different design principle from architecture based methods: \methodabbr designs a novel prompt-based memory to learn high-level \emph{instructions} from model inputs to steer model outputs and keeps the learned architecture fixed. In contrast, most architecture-based methods aim to separate model parameters.}

\zifeng{
Lastly, recent work of CTN~\cite{pham2020contextual} and DualNet~\cite{pham2021dualnet} start to consider knowledge management via a controller that models task-level information in addition to a backbone model. However, CTN still requires task identity at test time, while DualNet needs a rehearsal buffer to work. %
Moreover, CTN and DualNet are inspired from \final{a different perspective of} CLS, which suggests human beings achieve continual learning through two systems that facilitate fast learning and long-term remembering respectively. Interestingly, though we get our inspiration differently, \methodabbr could be interpreted through CLS theory exactly: The prompt pool deals with fast learning, and the backbone model serves as long-term memory.
}

\textbf{Prompting \zifeng{for transfer learning.}} 
The high-level idea of prompting is to apply a function to modify the input text, so that the language model gets additional information about the task. 
However, the design of a prompting function is challenging and requires heuristics. 
Recent work, including prompt tuning \cite{lester2021power} and prefix tuning \cite{li2021prefix}, seek to address this problem by applying learnable prompts in a continuous space, achieving excellent performance on transfer learning.
\zifeng{Prompts capture task-specific knowledge with much smaller additional parameters, than its competitors, such as Adapter~\cite{wang2020k, pfeiffer2020adapterfusion} and LoRA~\cite{hu2021lora}.} 
\zifeng{
The central idea of prompting is mainly designed for transfer learning. Note that
it is non-trivial to directly apply prompting in continual learning. Our proposed novel framework reveals its values to continual learning problems.
}

\section{Prerequisites}

\subsection{Continual learning protocols}
Continual learning is usually defined as training machine learning models on non-stationary data from sequential tasks. 
We define a sequence of tasks $\mathcal{D} = \{\mathcal{D}_1, \cdots, \mathcal{D}_T\}$, where the $t$-th task $\mathcal{D}_t=\{(\vx^t_i, y^t_i)\}_{i=1}^{n_t}$ contains tuples of the input sample $\vx^t_i \in \mathcal{X}$ and its corresponding label $y^t_i \in \mathcal{Y}$. %
The goal is to train a single model $f_\theta: \mathcal{X} \to \mathcal{Y}$ parameterized by $\theta$, such that it predicts the label $y = f_\theta(\vx) \in \mathcal{Y}$ given an unseen test sample $\vx$ from arbitrary tasks. 
Data from the previous tasks may not be seen anymore when training future tasks. 

Depending on the task transition environment, continual learning can be categorized into multiple settings with slightly different challenges.
The common task-, class-, and domain-incremental setting assume task data $\mathcal{D}_t$ arrives in sequence $t=\{1,..., T\}$ in a discrete manner. Different from class-incremental, task-incremental learning assumes task identity is known at test time \zifeng{and are often regarded as the simplest setting \cite{mai2021online, mehrabi2021survey}.}
Different from task- and class-incremental settings where each task has different classes, domain-incremental learning maintains the same set of classes for every task and only changes the distribution of $\vx$ by task. 
In the more challenging task-agnostic setting, task data in $\mathcal{D}$ changes smoothly, and the task identity $t$ is unknown. 
Our paper tackles the more challenging class-incremental and domain-incremental, and further explores the task-agnostic settings. 

\subsection{Prompt-based learning and baselines} \label{subsec:pt}
Prompt-based learning is an emerging technique in NLP. 
In contrast to traditional supervised fine-tuning, this type of methods design task-specific prompt functions to instruct pre-trained models perform corresponding tasks conditionally \cite{liu2021pre}.
One of recent techniques, Prompt Tuning (PT) \cite{lester2021power}, proposes to simply condition frozen T5-like language models \cite{raffel2020exploring} to perform down-stream NLP tasks by learning prompt parameters that are prepended to the input tokens to instruct the model prediction. 
Without loss of generality, here we introduce the definition of PT using the image modality transformer-based sequence models \cite{dosovitskiy2020image,vaswani2017attention}. 
The definition is easy to generalize to other modalities and sequence-based models. 

Given an input of 2D image $\vx \in \sR ^{H \times W \times C}$ and a pre-trained vision transformer (ViT) $f = f_r \circ f_e$ (excluding the classification head), where $f_e$ is the input embedding layer, and $f_r$ represents a stack of self-attention layers \cite{dosovitskiy2020image}. 
Images are reshaped to a sequence of flattened 2D patches $\vx_p \in \sR ^{L \times (S^2\cdot C)}$, where $L$ is the token length, \ie, the number of patches, $S$ is the patch size and $C$ is the original number of channels. 
To simplify notation, we assume the first token in $\vx_p$ is the \verb|[class]| token as part of the pre-trained model \cite{dosovitskiy2020image}. 
The pre-trained embedding layer $f_{e}: \sR ^{L \times (S^2\cdot C)} \to \sR ^{L \times D}$ projects the patched image to the embedding feature $\vx_e = f_{_e}(x) \in \sR ^{L \times D}$, where $D$ is the embedding dimension.
When solving multiple downstreaming tasks, we keep the large-scale pre-trained backbone frozen to maintain its generality following PT. 
The direct application of PT is to prepend learnable parameters $P_e \in \sR ^{L_p \times D}$, called a prompt, to the embedding feature $\vx_p = [P_e; \vx_e]$, and feed the extended sequences to the model function $f_r(\vx_p)$ for performing classification tasks. 
Different tasks have independent prompts and share one copy of the large model. 

Compared with ordinary fine-tuning, literature shows that prompt-based learning results in a sequence-based model having higher capacity to learn features \cite{liu2021pre,lester2021power}. \zifeng{
Despite its successes in transfer learning to train individual prompts for each task, prompting can not be directly applied to continual learning scenarios where test-time task identity is unknown.
}

\vspace{-.2cm}
\section{Learning to Prompt \zifeng{(L2P)}}

\subsection{From prompt to prompt pool} \label{sec:prompt_look_up}

The motivations of introducing prompt pool are threefold. 
First, the task identity at test time is unknown so training task-independent prompts is not feasible.
Second, even if the task-independent prompt can be known at test time, it prevents possible knowledge sharing between similar tasks \cite{hadsell2020embracing}. 
Third, while the naive way of learning a single shared prompt for all tasks enables knowledge sharing, it \zifeng{still causes severe forgetting issue} (see Section~\ref{sec:ablation}).
Ideally one would learn a model that is able to share knowledge when tasks are similar, while maintaining knowledge independent otherwise. 
Thus, we propose using a \emph{prompt pool} to store encoded knowledge, which can be flexibly grouped as an input to the model. 
The prompt pool is defined as 
\begin{equation}
\label{eq:prompt_pool}
    \pp=\{P_1, P_2, \cdots, P_M\}, \quad \text{$M = $ total $\#$ of prompts},
\end{equation}
where $P_j \in \sR^{L_p\times D}$ is a single prompt with token length $L_p$ and the same embedding size $D$ as $\vx_e$.
Following the notations in Section~\ref{subsec:pt}, we let $\vx$ and $\vx_e = f_e(\vx)$ be the input and its corresponding embedding feature, respectively. Note that we omit the task index $t$ of $\vx$ in our notation as our method is general enough to the task-agnostic setting. 
Denoting $\{s_i\}_{i=1}^{N}$ as a subset of $N$ indices from $[1, M]$, we can then adapt the input embedding as follows:
\begin{equation}
\label{eq:prompt_input}
    \vx_p = [P_{s_1}; \cdots; P_{s_N}; \vx_e], \quad 1 \leq N \leq M,
\end{equation}
where $;$ represents concatenation along the token length dimension.
Prompts are free to compose, so they can jointly encode knowledge (e.g. visual features or task \zifeng{information}) for the model to process. 
Ideally, we want to achieve a more fine-grained knowledge sharing scheme via prompt combinations at the instance-wise level: similar inputs tend to share more common prompts, and vice versa.

\subsection{Instance-wise prompt query} \label{sec:query}

We design a key-value pair based query strategy to dynamically select suitable prompts for different inputs (see Figure~\ref{fig:method}). 
This key-valued memory query mechanism shares some design principles with methods in other fields, such as Differentiable Neural Computer \cite{graves2016hybrid} and VQ-VAE \cite{oord2017neural}, which have external memory to maintain, and employ them for a different purpose.

We associate each prompt as value to a learnable key: $\{(\vk_1, P_1), (\vk_2, P_2), \cdots, (\vk_M, P_M)\}$, where $\vk_i \in \sR ^{D_k}$. \zifeng{And we denote the set of all keys by $\kk = \{\vk_i\}_{i=1}^{M}$.} Ideally, we would like to let the input instance itself decide which prompts to choose through query-key matching. To this end, we introduce a query function $q: \sR ^{H \times W \times C} \to \sR ^{D_k}$ that encodes input $\vx$ to the same dimension as the key.
Moreover, $q$ should be a deterministic function with respect to different tasks and has no learnable parameters. We directly use the whole pre-trained model as a frozen feature extractor to get the query features: $q(\vx) = f(\vx)[0,:]$ (we use the feature vector corresponding to \verb|[class]|). Other feature extractors like ConvNet are feasible as well. 

Denote $\gamma: \sR ^{D_k} \times \sR ^{D_k} \to \sR$ as a function to score the match between the query and prompt key (we find cosine distance works well).
Given an input $\vx$, we use $q(\vx)$ to lookup the top-$N$ keys by simply solving the objective:
\begin{equation}
\label{eq:lookup}
    \zifeng{\kk_\vx} = \underset{\{s_i\}_{i=1}^{N} \subseteq [1, M] }{\operatorname{argmin}} \quad \sum_{i=1}^{N} \gamma\left({q(\vx), \vk_{s_i}}\right),
\end{equation}
\zifeng{where $\kk_\vx$ represents the a subset of top-$N$ keys selected specifically for $\vx$ from $\kk$.}
Note that the design of this key-value strategy decouples the query mechanism learning and prompt learning processes, which has been experimentally shown to be critical (see Section~\ref{sec:ablation}).
Furthermore, querying prompts is done in an instance-wise fashion, which makes the whole framework \emph{task-agnostic}, meaning that the method works without needing clear task boundaries during training, nor task \zifeng{identity} at test time.

\textbf{Optionally diversifying prompt-selection.} 
Although our method does not need task boundary information, in real-world scenarios and experimental datasets, it is quite common that the task transition is discrete and so task boundaries are known at train time. 
We find that adding such a prior into our framework can help the model learn better task-specific prompts, especially when tasks have high diversity. 
To this end, we propose a simple extension to add task boundary prior, which is optional for \methodabbr.

During training of task $t$, we maintain a prompt frequency table $H_t = [h_1, h_2, \cdots, h_M]$, where each entry represents the normalized frequency of prompt $P_i$ being selected up until task $t-1$. 
To encourage the query mechanism to select diverse prompts, we modify \eqref{eq:lookup} to
\begin{equation}
\label{eq:lookup_v2}
    \zifeng{\kk_\vx} = \underset{\{s_i\}_{i=1}^{N} \subseteq [1, M] }{\operatorname{argmin}} \quad \sum_{i=1}^{N} \gamma\left({q(\vx), \vk_{s_i}}\right) \cdot h_{s_i},
\end{equation}
where $h_{s_i}$ penalizes the frequently-used prompts being selected to encourage diversified selection. 
Equation~\ref{eq:lookup_v2} is only applicable during training; at test time, \eqref{eq:lookup} is used.

\subsection{Optimization objective for \methodabbr}

\label{sec:training}
At every training step, after selecting $N$ prompts following the aforementioned query strategy, the adapted embedding feature $\vx_p$ is fed into the rest of the pre-trained model $f_r$ and the final classifier $g_{\phi}$ parametrized by $\phi$. 
Overall, we seek to minimize the end-to-end training loss function:
\begin{equation} \label{eq:full_loss}
\begin{split}
\underset{{\pp, \zifeng{\kk}, \phi}}{\operatorname{min}}\quad\mathcal{L}(g_{\phi} (f_r^{\text{avg}}(\vx_p)), y) + \lambda \sum_{\kk_{\vx}} \gamma\left({q(\vx), \vk_{s_i}}\right), \\ \text{\textit{s.t.}, \; \zifeng{$\kk_{\vx}$} is obtained with \eqref{eq:lookup}},
\end{split}
\end{equation}
where $f_r^{\text{avg}} = \text{AvgPool}(f_r(\vx_p)[\zifeng{0:NL_p},:])$, i.e., the output hidden vectors corresponding to the $N\cdot L_p$ prompt locations are averaged before the classification head.
The first term is the softmax cross-entropy loss, the second term is a surrogate loss to pull selected keys closer to corresponding query features. $\lambda$ is a scalar to weight the loss.

\begin{table*}[t]
\small
\caption{Results on class-incremental learning (i.e., task identity is unknown at test time). %
Compared methods are grouped based on \zifeng{different rehearsal buffer sizes. $0$ means no rehearsal is required,} where most SOTA methods are not applicable anymore.
\zizhaoz{Importantly, \methodabbr can attain competitive results without it and greatly outperform them with a small buffer size. }
} \vspace{-.3cm}
\label{table:task-inc}
\begin{center}
\begin{tabular}{l||c|cc||c|cc}
\toprule 
 \multirow{2}{*}{\textbf{Method}} & \multirow{2}{*}{\textbf{Buffer size}} & \multicolumn{2}{c||}{\textbf{Split CIFAR-100}} & \multirow{2}{*}{\textbf{Buffer size}} & \multicolumn{2}{c}{\textbf{5-datasets}} \\
& &  Average Acc ($\uparrow$) & Forgetting ($\downarrow$) & & Average Acc ($\uparrow$) & Forgetting ($\downarrow$) \\
\midrule
 
 FT-seq-frozen & \multirow{5}{*}{0} & 17.72\scriptsize{$\pm$0.34} & 59.09\scriptsize{$\pm$0.25} & \multirow{5}{*}{0} & 39.49\scriptsize{$\pm$0.12} & 42.62\scriptsize{$\pm$0.20} \\ 
 FT-seq & & 33.61\scriptsize{$\pm$0.85} & 86.87\scriptsize{$\pm$0.20} && 20.12\scriptsize{$\pm$0.42} & 94.63\scriptsize{$\pm$0.68} \\
 EWC \cite{kirkpatrick2017overcoming} & & 47.01\scriptsize{$\pm$0.29} & 33.27\scriptsize{$\pm$1.17} && 50.93\scriptsize{$\pm$0.09} & 34.94\scriptsize{$\pm$0.07} \\
 LwF \cite{li2017learning} & & 60.69\scriptsize{$\pm$0.63} & 27.77\scriptsize{$\pm$2.17} && 47.91\scriptsize{$\pm$0.33} & 38.01\scriptsize{$\pm$0.28} \\
 \textbf{\methodabbr (ours)} & & \bf{83.83\scriptsize{$\pm$0.04}} & \bf{7.63\scriptsize{$\pm$0.30}} && \bf{81.14 \scriptsize{$\pm$0.93}} & \bf{4.64 \scriptsize{$\pm$0.52}} \\
\midrule
   ER \cite{chaudhry2019tiny} & \multirow{6}{*}{10/class} & 67.87\scriptsize{$\pm$0.57} & 33.33\scriptsize{$\pm$1.28} & \multirow{6}{*}{5/class} & {80.32\scriptsize{$\pm$0.55}} & 15.69\scriptsize{$\pm$0.89} \\
  GDumb \cite{prabhu2020gdumb} & & 67.14\scriptsize{$\pm$0.37} & - && 56.99\scriptsize{$\pm$0.06} & - \\
 BiC \cite{wu2019large} & & 66.11\scriptsize{$\pm$1.76} & 35.24\scriptsize{$\pm$1.64} && 78.74\scriptsize{$\pm$1.41} & 21.15\scriptsize{$\pm$1.00} \\
 DER++ \cite{buzzega2020dark} & & 61.06\scriptsize{$\pm$0.87} & 39.87\scriptsize{$\pm$0.99} && 80.81\scriptsize{$\pm$0.07} & 14.38\scriptsize{$\pm$0.35} \\
 Co$^2$L \cite{cha2021co2l} & & 72.15\scriptsize{$\pm$1.32} & 28.55\scriptsize{$\pm$1.56} && 82.25\scriptsize{$\pm$1.17} & 17.52\scriptsize{$\pm$1.35} \\
 \textbf{\methodabbr-R (ours)} & & \bf{84.21\scriptsize{$\pm$0.53}} & \bf{7.72\scriptsize{$\pm$0.77}} && \bf{85.56\scriptsize{$\pm$0.95}} & \bf{4.22\scriptsize{$\pm$0.03}} \\
\midrule
  ER \cite{chaudhry2019tiny} &\multirow{6}{*}{50/class}& 82.53\scriptsize{$\pm$0.17} & 16.46\scriptsize{$\pm$0.25} &\multirow{6}{*}{10/class}& {84.26\scriptsize{$\pm$0.84}} & 12.85\scriptsize{$\pm$0.62} \\
 GDumb \cite{prabhu2020gdumb} & & 81.67\scriptsize{$\pm$0.02} & - && 70.76\scriptsize{$\pm$0.12} & - \\
 BiC \cite{wu2019large} & & 81.42\scriptsize{$\pm$0.85} & 17.31\scriptsize{$\pm$1.02} && 85.53\scriptsize{$\pm$2.06} & 10.27\scriptsize{$\pm$1.32} \\
 DER++ \cite{buzzega2020dark} & & 83.94\scriptsize{$\pm$0.34} & 14.55\scriptsize{$\pm$0.73} && 84.88\scriptsize{$\pm$0.57} & 10.46\scriptsize{$\pm$1.02} \\
 Co$^2$L \cite{cha2021co2l} & & 82.49\scriptsize{$\pm$0.89} & 17.48\scriptsize{$\pm$1.80} && 86.05\scriptsize{$\pm$1.03} & 12.28\scriptsize{$\pm$1.44} \\
 \textbf{\methodabbr-R (ours)} & & \bf{86.31\scriptsize{$\pm$0.59}} & \bf{5.83\scriptsize{$\pm$0.61}} && \bf{88.95\scriptsize{$\pm$0.78}} & \bf{4.92\scriptsize{$\pm$0.71}} \\
 \midrule
 Upper-bound & -& 90.85\scriptsize{$\pm$0.12} & - & - & 93.93\scriptsize{$\pm$0.18} & - \\
\bottomrule
\end{tabular}
\end{center}
\vspace{-.6cm}
\end{table*}

\begin{table}[t!]
\small
\caption{\cam{Class-incremental results on Split CIFAR-100 against architecture-based methods. \texttt{Diff = Upper-Bound Acc - Method Acc} (lower is better) measures how close the performance to the upper-bound of the used backbone.}}
\vspace{-6mm}
\label{table:architecture}
\begin{center}
\cam{
\begin{tabular}{l||c|lc}
\toprule 
 \multirow{2}{*}{\textbf{Method}} & \multirow{2}{*}{\textbf{Backbone}} & \multirow{2}{*}{\textbf{Avg. Acc ($\uparrow$)}} & \multirow{2}{*}{\textbf{Diff ($\downarrow$)}} \\
& &  &   \\
\midrule
  Upper-bound & \multirow{3}{*}{ResNet18}& 80.41 & -   \\
   SupSup~\cite{wortsman2020supermasks} & & 28.34\scriptsize{$\pm$2.45} & 52.07   \\
   DualNet~\cite{pham2021dualnet} & & 40.14\scriptsize{$\pm$1.64} & 40.27 \\
 \midrule
Upper-bound & \multirow{2}{*}{ViT-B/16}& 90.85 & -  \\
 \bf\methodabbr(ours) &  & \bf{83.83\scriptsize{$\pm$0.04}} & \bf 7.02 \\
 \bottomrule
\end{tabular}
}
\\
\end{center}
\vspace{-.7cm}
\end{table}

\section{Experiments} \label{sec:experiments}

To evaluate the proposed  \methodabbr, we closely follow the settings proposed in prior works \cite{lopez2017gradient,zeno2018task, van2019three}, and conduct comprehensive experiments. %
In particular, we mainly consider (1) the class-incremental setting, where the task identity is unknown during inference; (2) the domain-incremental setting, where the input domain shifts over time; (3) the task-agnostic setting, where there is no clear task boundary. 
\cam{We carefully compare \methodabbr with state-of-the-art (SOTA) methods of different categories under proper experiment settings.}
Moreover, we conduct extensive ablation studies to provide a deeper understanding of our method.

\subsection{Comparing methods} 
We compare \methodabbr against several baselines and state-of-the-art (SOTA) continual learning methods. \zizhaoz{Our method is based on a pre-trained ViT-B/16 \cite{vit,zhang2021aggregating}, which has become a common asset in advanced vision communities.
We carefully choose compared methods in the same environment for fair comparison.} \zifeng{Many recent methods claimed SOTA performance in the simplest task-incremental setting, where task identity is known at test time \cite{veniat2020efficient, ke2020continual, pham2020contextual}. 
We do not include these methods, since they are not applicable to more general class-incremental setting. %
We refer to multiple recent reviews papers \cite{delange2021continual, mai2021online} and recent work \cite{buzzega2020dark, prabhu2020gdumb, cha2021co2l} and select the most well-recognized and best-performing methods. For completeness, we also include naive sequential training approaches and representative regularization-based methods.}
\zifeng{Moreover, we refer to the original codebases for implementation and hyperparameter selection to ensure the best possible performance.
} 

\cam{\textbf{Baseline methods.}}
\zifeng{\emph{Upper-bound}} is the usual supervised finetuning on the i.i.d.~data of all tasks, which is the usually regarded as the upper bound performance a method can achieve. 
\emph{FT-seq-frozen} is the naive sequential fine-tuning approach with the pre-trained model frozen. 
\emph{FT-seq} instead fine-tunes pre-trained model weights as well. 
\zifeng{\emph{EWC} \cite{kirkpatrick2017overcoming} and \emph{LwF} \cite{li2017learning} are representative regularization-based approaches that are widely compared.}

\zifeng{\textbf{\cam{SOTA rehearsal-based methods.}}}
\zifeng{We select 5 advanced rehearsal-based methods to compare, including \emph{ER} \cite{chaudhry2019tiny, hayes2019memory}, \emph{GDumb} \cite{prabhu2020gdumb}, \emph{BiC} \cite{wu2019large}, \emph{DER++} \cite{buzzega2020dark} and \emph{Co$^2$L} \cite{cha2021co2l}. ER and GDumb are simple in concept, but they have achieved very strong performance not only in their own work, but in later literature \cite{mai2021online, buzzega2020dark} as well. DER++ and Co$^2$L are the latest SOTA methods.}

\cam{\textbf{SOTA architeture-based methods.} We select two representative architecture-based methods to compare. SupSup~\cite{wortsman2020supermasks} and DualNet~\cite{pham2021dualnet} are both based on ResNet18, recommended by their original authors. We compare the relative performance to the corresponding upper-bound performance for fairness.
}

\zifeng{\textbf{Our methods.} \emph{\methodabbr} is our proposed method without rehearsal buffer. \emph{\methodabbr-R} is \methodabbr equipped with a rehearsal buffer for a fair comparison with SOTA methods.}

\begin{table}
\small
\centering
\captionof{table}{Results on domain-incremental learning using CORe50 dataset, in terms of test accuracy. %
}
\vspace{-.2cm}
\label{table:domain-inc}
\begin{tabular}{l||c|c}
\toprule 
 \textbf{Method} & \textbf{Buffer size} & Test Acc ($\uparrow$)  \\
\toprule
EWC~\cite{kirkpatrick2017overcoming} & \multirow{3}{*}{0}  & 74.82\scriptsize{$\pm$0.60} \\
LwF~\cite{li2017learning} && 75.45\scriptsize{$\pm$0.40}  \\
\textbf{\methodabbr (ours)} && \bf{78.33\scriptsize{$\pm$0.06}} \\
\midrule
ER \cite{chaudhry2019tiny} &\multirow{6}{*}{50/class}& 80.10\scriptsize{$\pm$0.56}  \\
GDumb~\cite{prabhu2020gdumb} && 74.92\scriptsize{$\pm$0.25} \\
BiC~\cite{wu2019large} && 79.28\scriptsize{$\pm$0.30} \\
DER++~\cite{buzzega2020dark} && 79.70\scriptsize{$\pm$0.44} \\
Co$^2$L~\cite{cha2021co2l} && 79.75\scriptsize{$\pm$0.84} \\
\textbf{\methodabbr-R (ours)} && \bf{81.07\scriptsize{$\pm$0.13}} \\
\midrule
Upper-bound &-& 82.15\scriptsize{$\pm$0.37} \\
\bottomrule
\end{tabular}
\vspace{-5mm}
\end{table}

\begin{table}
\small
\captionof{table}{Results on task-agnostic continual learning using the Gaussian~scheduled CIFAR-100 dataset, in terms of test accuracy.
}
\vspace{-.2cm}
\label{table:task-agnostic}
\centering
\begin{tabular}{l||c|c}
\toprule 
 \textbf{Method} &\textbf{Buffer size}& Test Acc ($\uparrow$)  \\
\midrule
EWC~\cite{kirkpatrick2017overcoming} & \multirow{2}{*}{0} & 63.04\scriptsize{$\pm$0.42} \\
\textbf{\methodabbr (ours)} && \bf{88.34\scriptsize{$\pm$0.14}} \\
\midrule
ER~\cite{chaudhry2019tiny} &\multirow{4}{*}{50/class}& 82.63\scriptsize{$\pm$0.27}  \\
GDumb~\cite{prabhu2020gdumb} && 81.67\scriptsize{$\pm$0.02} \\
DER++~\cite{buzzega2020dark} && 85.24\scriptsize{$\pm$0.71} \\
\textbf{\methodabbr-R (ours)} && \bf{88.92\scriptsize{$\pm$0.39}} \\
\midrule
Upper-bound &-& 90.85\scriptsize{$\pm$0.12} \\
\bottomrule
\end{tabular}
\vspace{-.3cm}
\end{table}

\subsection{Datasets and experimental details}

\cam{
\textbf{Datasets.} We use Split CIFAR-100 \cite{krizhevsky2009learning} and 5-datasets \cite{ebrahimi2020adversarial} for class-incremental setting, CORe50 \cite{lomonaco2017core50} for domain-incremental setting, and Gaussian scheduled CIFAR-100 \cite{shanahan2021encoders} for task-agnostic setting, to evaluate the effectiveness of our method. Details of the datasets are introduced in Appendix~\ref{appendix:dataset}.
}

\textbf{Evaluation metrics.} For settings with task boundaries and where each task has an associated test set, we use two metrics, \emph{Average accuracy} (higher is better) and \emph{Forgetting} (lower is better), which are widely used in previous works \cite{lopez2017gradient, chaudhry2018efficient, mai2021online}. 
For settings without task boundary or where there is only a single test set available, we report the final test accuracy following the common protocol~\cite{lomonaco2017core50, shanahan2021encoders}.

\begin{table}[t]
\small
\caption{Ablation studies. See text for detailed explanations. %
} \vspace{-.5cm}
\label{table:ablation_5_datasets}
\begin{center}
\begin{tabular}{l|cc}
\toprule  
\multirow{2}{*}{\textbf{Ablated components}} & \multicolumn{2}{c}{\textbf{5-datasets}} \\
 & Average Acc ($\uparrow$) & Forgetting ($\downarrow$) \\
\midrule
 w/o prompt pool &  51.96 & 26.60 \\
 w/o key-value pair &  58.33 & 20.45 \\
 w/o diversified selection  & 62.26 & 17.84 \\ \midrule
 none & \bf{81.14} & \bf{4.64} \\
\bottomrule
\end{tabular}
\end{center}
\vspace{-.5cm}
\end{table}

\textbf{Training details.} 
For \methodabbr, we train all models using Adam \cite{kingma2014adam} with $\beta_1= 0.9$ and $\beta_2=0.999$, a batch size of 128, and a constant learning rate of $0.03$ for all settings. 
Input images are resized to $224\times224$ and normalized to the range of $[0, 1]$ to match the pretraining setting. 
As pointed out by \cite{buzzega2020dark}, training multiple epochs for each task disentangles the effects of possible underfitting from forgetting. 
Thus, we train every task for 5 epochs in the class- and domain-incremental settings. 
However, in the task-agnostic setting where we don't have the concept of a task, we follow \cite{shanahan2021encoders} to train every batch only once. 
We set $M= 10,N= 5,L_p= 5$ for all CIFAR-100 based datasets and CORe50. 
For 5-datasets, we use $M= 20,N= 4,L_p= 5$. 
Prompts only add $46,080$ and $92,160$ parameters to the original pre-trained model for these two settings, leading to a small $0.05\%$ and $ 0.11\%$ total parameter increase, respectively. \zizhaoz{We apply the optional prompt selection strategy introduced in~\ref{sec:query} to this dataset.}
We find $\lambda$ in \eqref{eq:full_loss} is not sensitive and works well in a large range, so we set $\lambda=0.5$ consistently for all datasets. \zifeng{Main experimental results are averaged over 3 runs, and corresponding standard deviation is reported as well.}

\begin{figure}[t]
\centering
\begin{tabular}{c}
   \includegraphics[width=0.99\columnwidth]{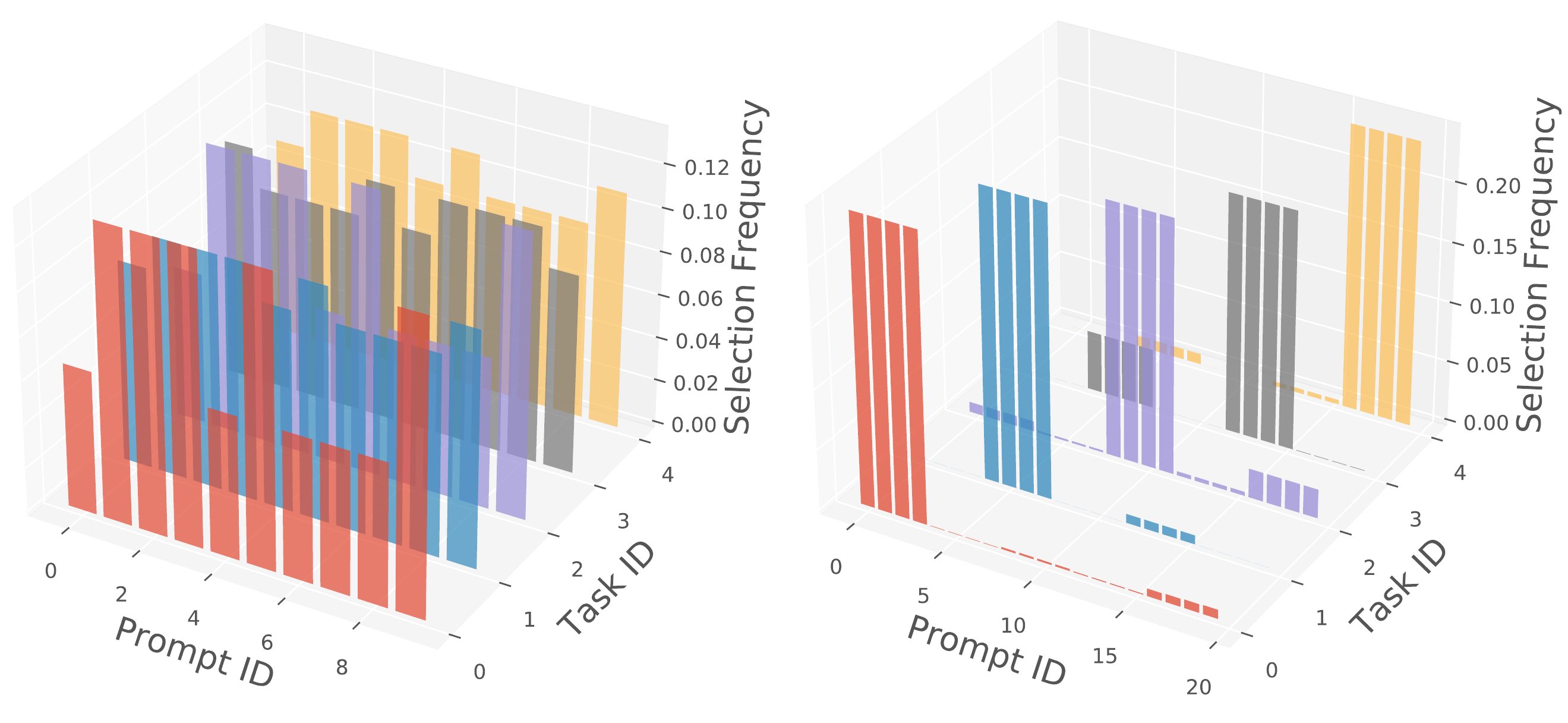}
\end{tabular}
\vspace{-.2cm}
\caption{Prompt selection histograms for (left) Split CIFAR-100 and (right) 5-datasets. \finalx{CIFAR-100 has higher intra-task similarity compared to 5-datasets, thus largely sharing prompts between tasks results in good performance, while 5-datasets favors more task-specific prompts.} We only show the first 5 tasks for Split CIFAR-100 for better readability.}
	\label{fig:prompt_hist} \vspace{-3mm}
\end{figure}

\begin{figure*}[t]
\centering
\begin{tabular}{c c c}
   \centering
   \includegraphics[width=0.45\columnwidth]{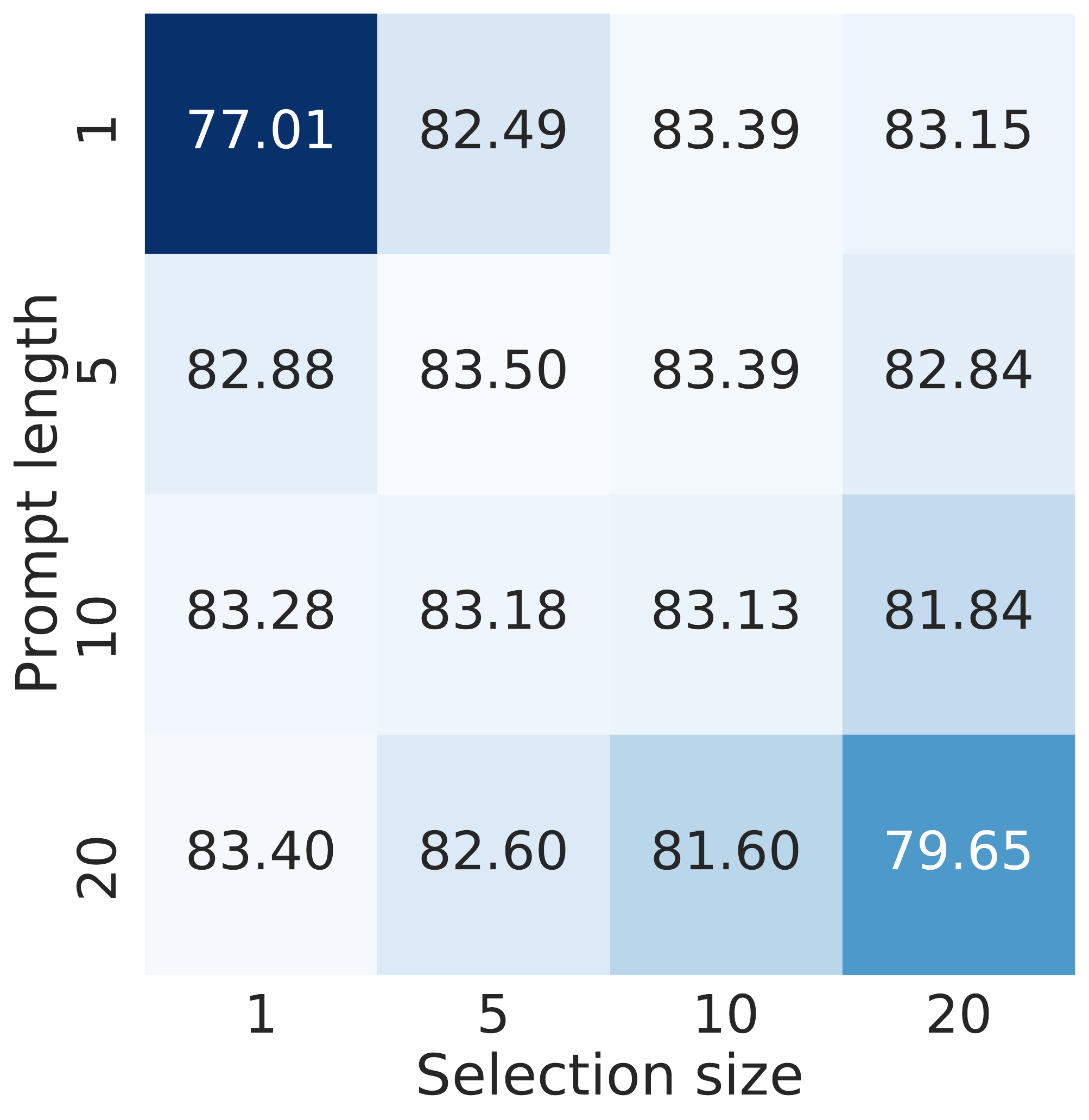} &
   \includegraphics[width=0.45\columnwidth]{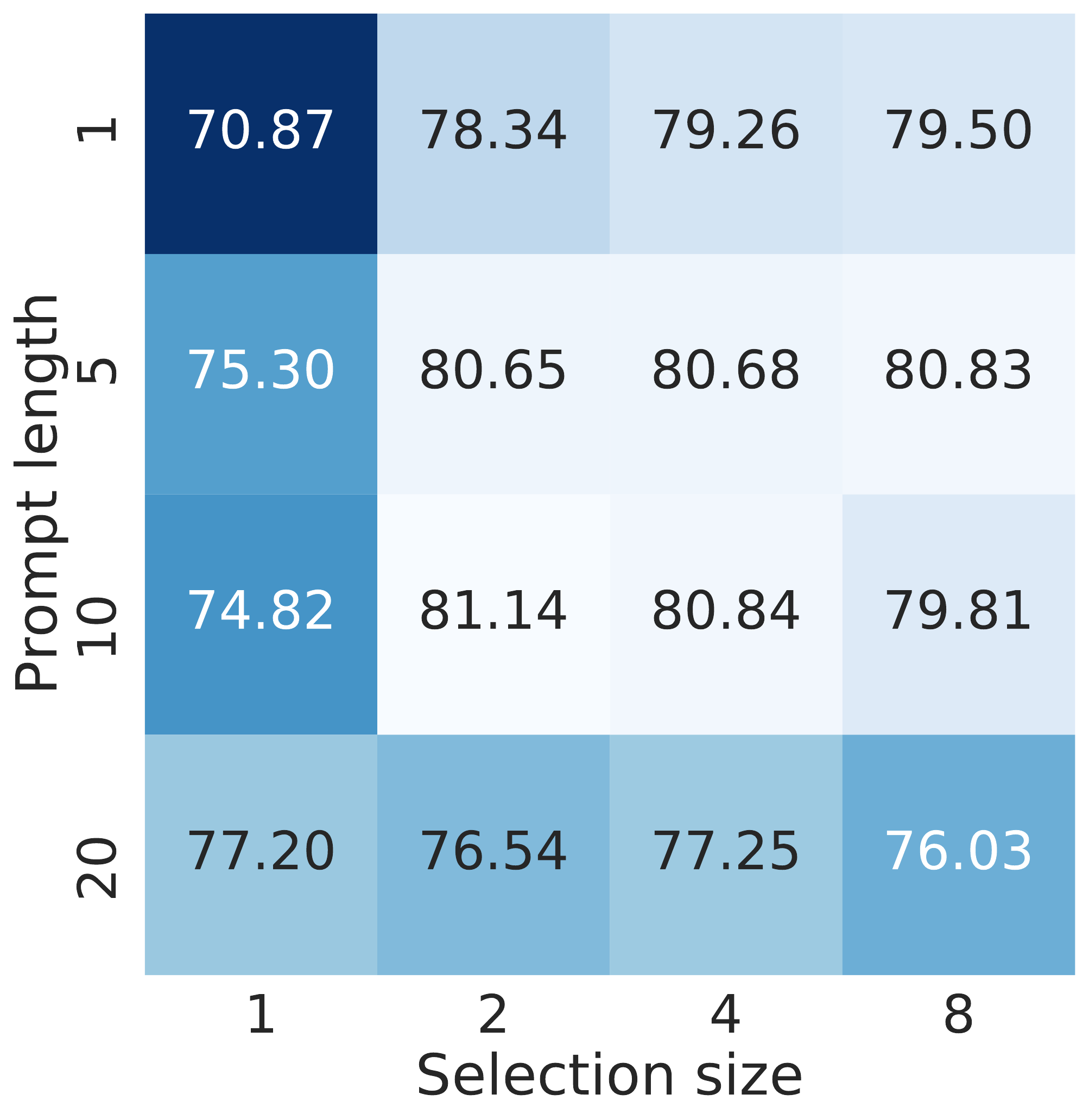} &
   \includegraphics[width=0.45\columnwidth]{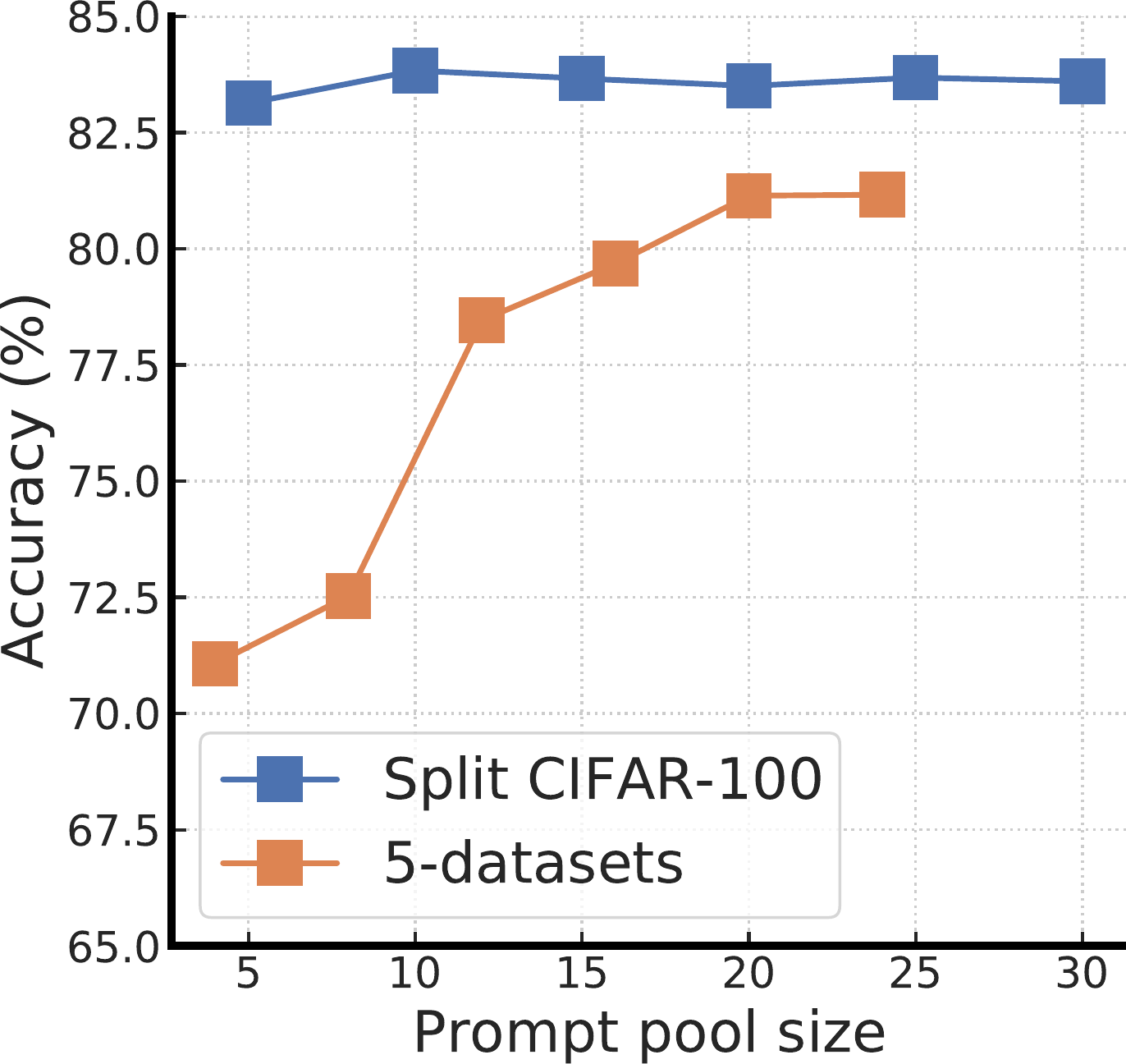} \\
\end{tabular} 
\vspace{-.3cm}
\caption{Left-Middle: Average accuracy w.r.t prompt length $L_p$ and prompt selection size $N$ for Split CIFAR-100 and 5-datasets, respectively, given $M=20$. Right: Average accuracy (\%) w.r.t. prompt pool size $M$, given $L_p=5$, $N=5$ for Split CIFAR-100 and $L_p=5$, $N=4$ for 5-datasets. }  \label{fig:ablation_N_L} 
\vspace{-.4cm}
\end{figure*}

\subsection{Main results}

\textbf{Results on class-incremental learning.} Table~\ref{table:task-inc} summarizes the results on these two class-incremental benchmarks.
\zifeng{\methodabbr outperforms all comparing methods consistently under different configurations, in terms of both average accuracy and forgetting. We observe that when the buffer size is relatively large, \methodabbr not only outperforms all other methods, but also closes a significant part of the gap to the upper bound performance under the i.i.d.~setting. 
When the buffer size gets smaller, \methodabbr outperforms others by a even larger margin. Finally, when there is no buffer, rehearsal-based methods are no longer capable, while \methodabbr still remains superior performance by beating regularization-based methods, and outperforms almost all rehearsal-based methods when buffer is small. %
}

\cam{Table~\ref{table:architecture} shows the comparison between \methodabbr and architecture-based methods on Split CIFAR-100. Instead of absolute performance in average accuracy, we use difference to upper-bound (Diff) to measure the performance of each method given a specific architecture. We observe that \methodabbr outperforms both methods with (DualNet) or without (SupSup) rehearsal buffer, by a large margin. %
}

\zifeng{
The outstanding performance of \methodabbr \cam{over all competing methods} indicates that our proposed prompt pool successfully accumulates knowledge from experiences, thus it can overall improve the learning performance while mitigating catastrophic forgetting even without a rehearsal buffer. 
}

\zifeng{\textbf{Results on domain-incremental learning.} Table~\ref{table:domain-inc} summarizes the results on the domain-incremental setting. 
\methodabbr remains the best performance compared with other methods. %
Interestingly, all rehearsal-based comparing methods perform quite closely (except GDumb). %
\zifeng{The observation of relatively modest performance gap between baseline methods and the upper-bound result has also been reported in \cite{lomonaco2017core50}, thus there is indeed a significant performance gap between our method and others.}
}

\zifeng{
\textbf{Results on task-agnostic learning.}
Although task-agnostic setting is usually considered more challenging \cite{shanahan2021encoders}, the topic is  under-investigated. 
We conduct more exploratory studies on task-agnostic settings.
Table~\ref{table:task-agnostic} summarizes the results on the challenging task-agnostic learning setting. We do not compare with LwF, BiC and Co$^2$L \zizhaoz{since they require task boundary to save model snapshots and calculate distillation loss. It is beyond our scope to extend them to this setting. We also use the online version of EWC proposed by \cite{chaudhry2018riemannian} for the task agnostic setting.}
\zizhaoz{Since all compared methods are based on pre-trained models, the absolute numbers are not too far away from Upper-bound.} As can been seen, rehearsal-based methods have clear advantages.
Nevertheless, \methodabbr still achieves the best performance even when buffer size is zero, among all methods, including ones have a rehearsal buffer. 
We believe that the smoother transition of tasks implicitly help \methodabbr consolidate knowledge into prompts. Since we have better prompts, the benefit of a rehearsal buffer is naturally weakened. 
}

\subsection{Effectiveness of core designs} \label{sec:ablation}

\textbf{Effect of prompt related components for \methodabbr.}
Table~\ref{table:ablation_5_datasets} (row 1) removes the prompt pool design and uses a single prompt to train sequentially. The performance has a significant drop, suggesting that \finalx{a single prompt suffers severe catastrophic forgetting and knowledge interference between tasks, while our design of} prompt pool encodes task-invariant and task-specific knowledge well. Table~\ref{table:ablation_5_datasets} (row 2) removes the learnable key associated with prompts and directly uses mean of prompts as keys. %
As results show, learnable keys play an important role to decouple the query and prompt learning processes. Table~\ref{table:ablation_5_datasets} (row 3) removes the diversified prompt selection (only used in 5-dataset experiments). Basically, removing it allows instances from different tasks to choose prompts freely. The decrease in performance suggests that, when tasks are diverse, adding this strategy indeed reduces unnecessary knowledge sharing and thus mitigating interference between unrelated tasks.

To better understand the prompt selection mechanism, we plot the prompt selection histograms for each task in both Split CIFAR-100 and 5-datasets in Figure~\ref{fig:prompt_hist} under the best-performing parameters settings, respectively.
From the plot of Split CIFAR-100 (left), the tasks largely share all prompts, meaning that our prompt selection mechanism encourages more knowledge sharing between similar tasks. 
In contrast, in the plot of 5-datasets (right), diverse tasks require more task-specific prompts and share less.

\textbf{Effect of hyperparameters for \methodabbr.} 
Recall that there are three key hyperparameters, including the size of the prompt pool $M$, length of a single prompt $L_p$, and the selection size $N$ used as model input. 
Intuitively, $M$ decides the total capacity of learnable prompts. 
$L_p$ decides capacity of a singe prompt (which jointly encodes certain knowledge), and $L_p \times N$ decides the total size used to prepend the input.
From the results on both datasets (Figure~\ref{fig:ablation_N_L} (left-middle)), a too small $L_p$ always negatively affects results, \finalx{while an oversized prompt may introduce knowledge underfitting.}
We hypothesize that a reasonable capacity of a single prompt is critical to encode a certain aspect of shared knowledge. 
Increasing the prompt pool size shows positive effect on performance as shown in Figure~\ref{fig:ablation_N_L} (right) on 5-datasets \finalx{while not as effective on Split CIFAR-100}, suggesting a large enough pool size is needed to encode task-specific knowledge when tasks are diverse.

\section{Conclusion}
This paper presents a novel method to address some of the key challenges in continual learning with a method that can achieve strong performance without a need for rehearsal and task identity. \methodabbr introduces prompt-based learning to continual learning and proposes a novel technique to enable a single pre-trained model to adapt to sequential tasks via a shared prompt pool, successfully mitigating the catastrophic forgetting problem.
The resulting method significantly outperforms previous SOTA on several continual learning problems, including class-incremental and domain-incremental. We show our method is general enough to handle even more challenging task-agnostic settings where previous methods are incapable of. 

\section*{Acknowledgments}
We would like to thank Chun-Liang Li, Jeremy Martin Kubica, Sayna Ebrahimi, Stratis Ioannidis, Nan Hua, and Emmanouil Koukoumidis, for their valuable discussions.

\newpage
\onecolumn
\begin{alphasection}
\section{Potential negative societal impact}
\methodabbr is a strong continual learning method and has great potential to be applied in various fields.  
However, there are some ways it could be misused.
Our method takes a well-pretrained model as a backbone, thus any bias and fairness issues \cite{mehrabi2021survey} in the original model may be carried over during the continual learning process. 
We encourage any users to thoroughly check the pretrained model to mitigate any bias and fairness issues. 
Moreover, the method could be deployed in safety-critical applications, such as autonomous driving systems \cite{grigorescu2020survey}, which may present potential security issues in terms of adversarial attacks \cite{madry2017towards}. 
We would recommend testing the robustness of our method in future work and design corresponding defense techniques to deal with potential security concerns.

\section{Limitations}
Although our method is demonstrated on vision models, it does not make any assumption of modalities. 
We leave exploration on other modalities as future work. Additionally, \methodabbr assumes there are pre-trained sequence-based models.  While they have become common assets and future directions in advanced communities, how to generalize our framework to other vision architectures (e.g. ConvNet) could be an appealing research direction.

How to achieve continual learning that can satisfy the real-world requirements is an important direction that remains challenging. For example, the task-agnostic setting is known as the most challenging setting and is very close to real-world scenarios. Although our method takes a step further towards this goal, however, the current 
commonly used Gaussian scheduled CIFAR-100 is synthetic and still far from realistic. Thus, we think it also requires more complex benchmarks to evaluate the ability of task-agnostic continual learning methods and push forward the advances of this real-world challenge.

\section{Dataset details and licensing information} \label{appendix:dataset}
\textbf{Split CIFAR-100 (class-incremental).} This dataset splits the original CIFAR-100 \cite{krizhevsky2009learning} into 10 tasks, 10 disjoint classes per task. Since the tasks are from a single original dataset, they share some similarities and some classes could be from the same superclass. 
\zizhaoz{Although CIFAR-100 is a simple image classification dataset, it remains quite challenging for continual learning studies, especially in the class-incremental setting \cite{mai2021online}.}

\textbf{5-datasets (class-incremental).} We also use a challenging dataset proposed in \cite{ebrahimi2020adversarial}. This dataset consists of five image classification datasets: CIFAR-10, MNIST~\cite{lecun1998mnist}, Fashion-MNIST~\cite{xiao2017fashion}, SVHN~\cite{netzer2011reading}, and notMNIST~\cite{notmnist}. Although each dataset alone is not hard, the sequential training of them is fairly challenging even with ImageNet pre-trained models, since models are susceptible to forgetting when the tasks are diverse \cite{mehtaempirical}.

\textbf{CORe50 (domain-incremental)}. 
This is a widely used dataset specifically designed for continual object recognition \cite{lomonaco2017core50}. 
It is a collection of 50 objects collected in 11 distinct domains, where 8 of them (120,000 samples) are used for training, and the rest are considered as a single test set (45,000). %
Methods are trained on each domain sequentially.

\textbf{Gaussian scheduled CIFAR-100 (task-agnostic).} 
The distribution of data shifts gradually throughout the learning process \cite{shanahan2021encoders}, the probability that a class is present in a batch follows a Gaussian distribution centered with intervals. There is no explicit task boundaries between batches, thus requiring methods to be able to implicitly adapt to non-stationary data distribution without utilizing any task-specific information during both training and inference.

\begin{itemize}
    \item CIFAR-10 and CIFAR-100 \cite{krizhevsky2009learning}, Fashion-MNIST \cite{xiao2017fashion} are licensed under the MIT license.
    \item MNIST \cite{lecun1998mnist} is licensed under the Creative Commons Attribution-Share Alike 3.0 license.
    \item CORe50 \cite{lomonaco2017core50} is under the Creative Commons Attribution 4.0 International license.
    \item The licensing information is not available for SVHN\cite{netzer2011reading}, notMNIST \cite{notmnist}.
\end{itemize}

\begin{algorithm}[t]
\SetAlgoLined
\textbf{Input:} Pre-trained input embedding layer $f_e$, pre-trained self-attention layers $f_r$, final classification layer $g_{\phi}$, number of tasks $T$, training set ${\{(\vx^t_i, y^t_i)\}_{i=1}^{n_t}}\}_{t=1}^{T}$, prompt pool $\pp=\{P_j\}_{j=1}^{M}$, prompt keys $\kk=\{K_j\}_{j=1}^{M}$, number of training epochs of the $t$-th task $E_t$, learning rate $\eta$, balancing parameter $\lambda$\\
\textbf{Initialize:} $g_{\phi},\ \pp,\ \kk$\\
\For{$t = 1,\cdots,T$}{
    \For{$e = 1, \cdots, E_t$}{
      Draw a mini-batch $B=\{ (\vx_i^t, y_i^t)\}_{i=1}^{l}$ \\
      Initialize the sets of chosen keys and prompts for current batch: $\kk_B = \{\}, \pp_B = \{\}$.\\
      \For{$(\vx, y)$ in $B$}{
        Calculate query feature $q(\vx)$\\
        Lookup top-$N$ keys by solving $\kk_\vx = \underset{\{s_i\}_{i=1}^{N} \subseteq [1, M] }{\operatorname{argmin}} \sum_{i=1}^{N} \gamma\left({q(\vx), \vk_{s_i}}\right)$ (equation~3) \\
        Select top-$N$ prompts associated with the keys in $\pp_{\vx}$ \\
        Calculate the input embedding sequence $\vx_e = f_e(\vx)$ \\
        Prepending $\vx_e$ with corresponding top-$N$ prompts by $\vx_p = [P_{s_1}; \cdots; P_{s_N}; \vx_e]$ (equation~2) \\
        Calculate per sample loss $\mathcal{L}_{\vx} = \mathcal{L}(g_{\phi} (f_r^{\text{avg}}(\vx_p)), y) + \lambda \sum_{\kk_{\vx}} \gamma\left({q(\vx), \vk_{s_i}}\right)$ (equation~5) \\
        Update sets of chosen keys and prompts: $\kk_B = \kk_B \cup \kk_{\vx}, \pp_B = \pp_B \cup \pp_{\vx}$
      }
      Calculate per batch loss $\mathcal{L}_{B}$ by accumulating $\mathcal{L}_{\vx}$ \\
      \For{$(K, P)$ in $\operatorname{zip}(\kk_B, \pp_B)$}{
        Update $K$ by $K \leftarrow K - \eta \nabla_{K} \mathcal{L}_{B}$ \\
        Update $P$ by $P \leftarrow P - \eta \nabla_{P} \mathcal{L}_{B}$ \\
      }
      Update $\phi$ by $\phi \leftarrow \phi - \eta \nabla_{\phi} \mathcal{L}_{B}$ \\
    }
 }
 \caption{\method (\methodabbr)}
 \label{alg:l2p}
\end{algorithm}

\section{Algorithm details}
To better illustrate our proposed method, we present a whole picture of the training procedure in Algorithm~\ref{alg:l2p}. Note that for prediction, we simply replace loss calculation to label prediction. Optionally, we can replace the top-$N$ keys lookup by equation~4, when task boundary prior is known.

\end{alphasection}


\begin{thebibliography}{10}\itemsep=-1pt

\bibitem{aljundi2018memory}
Rahaf Aljundi, Francesca Babiloni, Mohamed Elhoseiny, Marcus Rohrbach, and
  Tinne Tuytelaars.
\newblock Memory aware synapses: Learning what (not) to forget.
\newblock In {\em ECCV}, 2018.

\bibitem{notmnist}
Yaroslav Bulatov.
\newblock notmnist dataset, 2011.

\bibitem{buzzega2020dark}
Pietro Buzzega, Matteo Boschini, Angelo Porrello, Davide Abati, and Simone
  Calderara.
\newblock Dark experience for general continual learning: a strong, simple
  baseline.
\newblock In {\em NeurIPS}, 2020.

\bibitem{cha2021co2l}
Hyuntak Cha, Jaeho Lee, and Jinwoo Shin.
\newblock Co2l: Contrastive continual learning.
\newblock In {\em ICCV}, 2021.

\bibitem{chaudhry2018riemannian}
Arslan Chaudhry, Puneet~K Dokania, Thalaiyasingam Ajanthan, and Philip~HS Torr.
\newblock Riemannian walk for incremental learning: Understanding forgetting
  and intransigence.
\newblock In {\em ECCV}, pages 532--547, 2018.

\bibitem{chaudhry2020using}
Arslan Chaudhry, Albert Gordo, Puneet~Kumar Dokania, Philip Torr, and David
  Lopez-Paz.
\newblock Using hindsight to anchor past knowledge in continual learning.
\newblock {\em arXiv preprint arXiv:2002.08165}, 2(7), 2020.

\bibitem{chaudhry2018efficient}
Arslan Chaudhry, Marc'Aurelio Ranzato, Marcus Rohrbach, and Mohamed Elhoseiny.
\newblock Efficient lifelong learning with a-gem.
\newblock {\em arXiv preprint arXiv:1812.00420}, 2018.

\bibitem{chaudhry2019tiny}
Arslan Chaudhry, Marcus Rohrbach, Mohamed Elhoseiny, Thalaiyasingam Ajanthan,
  Puneet~K Dokania, Philip~HS Torr, and Marc'Aurelio Ranzato.
\newblock On tiny episodic memories in continual learning.
\newblock {\em arXiv preprint arXiv:1902.10486}, 2019.

\bibitem{delange2021continual}
Matthias Delange, Rahaf Aljundi, Marc Masana, Sarah Parisot, Xu Jia, Ales
  Leonardis, Greg Slabaugh, and Tinne Tuytelaars.
\newblock A continual learning survey: Defying forgetting in classification
  tasks.
\newblock {\em TPAMI}, 2021.

\bibitem{dosovitskiy2020image}
Alexey Dosovitskiy, Lucas Beyer, Alexander Kolesnikov, Dirk Weissenborn,
  Xiaohua Zhai, Thomas Unterthiner, Mostafa Dehghani, Matthias Minderer, Georg
  Heigold, Sylvain Gelly, et~al.
\newblock An image is worth 16x16 words: Transformers for image recognition at
  scale.
\newblock {\em ICLR}, 2021.

\bibitem{vit}
Alexey Dosovitskiy, Lucas Beyer, Alexander Kolesnikov, Dirk Weissenborn,
  Xiaohua Zhai, Thomas Unterthiner, Mostafa Dehghani, Matthias Minderer, Georg
  Heigold, Sylvain Gelly, Jakob Uszkoreit, and Neil Houlsby.
\newblock An image is worth 16x16 words: Transformers for image recognition at
  scale.
\newblock In {\em ICLR}. OpenReview.net, 2021.

\bibitem{ebrahimi2020adversarial}
Sayna Ebrahimi, Franziska Meier, Roberto Calandra, Trevor Darrell, and Marcus
  Rohrbach.
\newblock Adversarial continual learning.
\newblock In {\em ECCV}, 2020.

\bibitem{farquhar2018towards}
Sebastian Farquhar and Yarin Gal.
\newblock Towards robust evaluations of continual learning.
\newblock {\em arXiv preprint arXiv:1805.09733}, 2018.

\bibitem{graves2016hybrid}
Alex Graves, Greg Wayne, Malcolm Reynolds, Tim Harley, Ivo Danihelka, Agnieszka
  Grabska-Barwi{\'n}ska, Sergio~G{\'o}mez Colmenarejo, Edward Grefenstette,
  Tiago Ramalho, John Agapiou, et~al.
\newblock Hybrid computing using a neural network with dynamic external memory.
\newblock {\em Nature}, 538(7626):471--476, 2016.

\bibitem{grigorescu2020survey}
Sorin Grigorescu, Bogdan Trasnea, Tiberiu Cocias, and Gigel Macesanu.
\newblock A survey of deep learning techniques for autonomous driving.
\newblock {\em Journal of Field Robotics}, 37(3):362--386, 2020.

\bibitem{hadsell2020embracing}
Raia Hadsell, Dushyant Rao, Andrei~A Rusu, and Razvan Pascanu.
\newblock Embracing change: Continual learning in deep neural networks.
\newblock {\em Trends in cognitive sciences}, 2020.

\bibitem{hayes2019memory}
Tyler~L Hayes, Nathan~D Cahill, and Christopher Kanan.
\newblock Memory efficient experience replay for streaming learning.
\newblock In {\em ICRA}, 2019.

\bibitem{hu2021lora}
Edward~J Hu, Yelong Shen, Phillip Wallis, Zeyuan Allen-Zhu, Yuanzhi Li, Shean
  Wang, Lu Wang, and Weizhu Chen.
\newblock Lora: Low-rank adaptation of large language models.
\newblock {\em arXiv preprint arXiv:2106.09685}, 2021.

\bibitem{ke2020continual}
Zixuan Ke, Bing Liu, and Xingchang Huang.
\newblock Continual learning of a mixed sequence of similar and dissimilar
  tasks.
\newblock {\em NeurIPS}, 33, 2020.

\bibitem{kingma2014adam}
Diederik~P Kingma and Jimmy Ba.
\newblock Adam: A method for stochastic optimization.
\newblock {\em arXiv preprint arXiv:1412.6980}, 2014.

\bibitem{kirkpatrick2017overcoming}
James Kirkpatrick, Razvan Pascanu, Neil Rabinowitz, Joel Veness, Guillaume
  Desjardins, Andrei~A Rusu, Kieran Milan, John Quan, Tiago Ramalho, Agnieszka
  Grabska-Barwinska, et~al.
\newblock Overcoming catastrophic forgetting in neural networks.
\newblock {\em PNAS}, 114(13):3521--3526, 2017.

\bibitem{krizhevsky2009learning}
Alex Krizhevsky, Geoffrey Hinton, et~al.
\newblock Learning multiple layers of features from tiny images.
\newblock 2009.

\bibitem{kumaran2016learning}
Dharshan Kumaran, Demis Hassabis, and James~L McClelland.
\newblock What learning systems do intelligent agents need? complementary
  learning systems theory updated.
\newblock {\em Trends in cognitive sciences}, 20(7):512--534, 2016.

\bibitem{lecun1998mnist}
Yann LeCun.
\newblock The mnist database of handwritten digits.
\newblock {\em http://yann. lecun. com/exdb/mnist/}, 1998.

\bibitem{lester2021power}
Brian Lester, Rami Al-Rfou, and Noah Constant.
\newblock The power of scale for parameter-efficient prompt tuning.
\newblock {\em arXiv preprint arXiv:2104.08691}, 2021.

\bibitem{li2019learn}
Xilai Li, Yingbo Zhou, Tianfu Wu, Richard Socher, and Caiming Xiong.
\newblock Learn to grow: A continual structure learning framework for
  overcoming catastrophic forgetting.
\newblock In {\em ICML}, pages 3925--3934. PMLR, 2019.

\bibitem{li2021prefix}
Xiang~Lisa Li and Percy Liang.
\newblock Prefix-tuning: Optimizing continuous prompts for generation.
\newblock {\em arXiv preprint arXiv:2101.00190}, 2021.

\bibitem{li2017learning}
Zhizhong Li and Derek Hoiem.
\newblock Learning without forgetting.
\newblock {\em TPAMI}, 40(12):2935--2947, 2017.

\bibitem{liu2021pre}
Pengfei Liu, Weizhe Yuan, Jinlan Fu, Zhengbao Jiang, Hiroaki Hayashi, and
  Graham Neubig.
\newblock Pre-train, prompt, and predict: A systematic survey of prompting
  methods in natural language processing.
\newblock {\em arXiv preprint arXiv:2107.13586}, 2021.

\bibitem{lomonaco2017core50}
Vincenzo Lomonaco and Davide Maltoni.
\newblock Core50: a new dataset and benchmark for continuous object
  recognition.
\newblock In {\em Conference on Robot Learning}, 2017.

\bibitem{loo2020generalized}
Noel Loo, Siddharth Swaroop, and Richard~E Turner.
\newblock Generalized variational continual learning.
\newblock {\em arXiv preprint arXiv:2011.12328}, 2020.

\bibitem{lopez2017gradient}
David Lopez-Paz and Marc'Aurelio Ranzato.
\newblock Gradient episodic memory for continual learning.
\newblock {\em NeurIPS}, 2017.

\bibitem{madry2017towards}
Aleksander Madry, Aleksandar Makelov, Ludwig Schmidt, Dimitris Tsipras, and
  Adrian Vladu.
\newblock Towards deep learning models resistant to adversarial attacks.
\newblock {\em arXiv preprint arXiv:1706.06083}, 2017.

\bibitem{mai2021online}
Zheda Mai, Ruiwen Li, Jihwan Jeong, David Quispe, Hyunwoo Kim, and Scott
  Sanner.
\newblock Online continual learning in image classification: An empirical
  survey.
\newblock {\em arXiv preprint arXiv:2101.10423}, 2021.

\bibitem{mallya2018packnet}
Arun Mallya and Svetlana Lazebnik.
\newblock Packnet: Adding multiple tasks to a single network by iterative
  pruning.
\newblock In {\em CVPR}, 2018.

\bibitem{mcclelland1995there}
James~L McClelland, Bruce~L McNaughton, and Randall~C O'Reilly.
\newblock Why there are complementary learning systems in the hippocampus and
  neocortex: insights from the successes and failures of connectionist models
  of learning and memory.
\newblock {\em Psychological review}, 102(3):419, 1995.

\bibitem{mccloskey1989catastrophic}
Michael McCloskey and Neal~J Cohen.
\newblock Catastrophic interference in connectionist networks: The sequential
  learning problem.
\newblock In {\em Psychology of learning and motivation}, volume~24, pages
  109--165. Elsevier, 1989.

\bibitem{mehrabi2021survey}
Ninareh Mehrabi, Fred Morstatter, Nripsuta Saxena, Kristina Lerman, and Aram
  Galstyan.
\newblock A survey on bias and fairness in machine learning.
\newblock {\em ACM Computing Surveys (CSUR)}, 54(6):1--35, 2021.

\bibitem{mehtaempirical}
Sanket~Vaibhav Mehta, Darshan Patil, Sarath Chandar, and Emma Strubell.
\newblock An empirical investigation of the role of pre-training in lifelong
  learning.
\newblock {\em ICML Workshop on Theory and Foundation of Continual Learning},
  2021.

\bibitem{netzer2011reading}
Yuval Netzer, Tao Wang, Adam Coates, Alessandro Bissacco, Bo Wu, and Andrew~Y
  Ng.
\newblock Reading digits in natural images with unsupervised feature learning.
\newblock In {\em NIPS}, 2011.

\bibitem{oord2017neural}
Aaron van~den Oord, Oriol Vinyals, and Koray Kavukcuoglu.
\newblock Neural discrete representation learning.
\newblock {\em arXiv preprint arXiv:1711.00937}, 2017.

\bibitem{parisi2019continual}
German~I Parisi, Ronald Kemker, Jose~L Part, Christopher Kanan, and Stefan
  Wermter.
\newblock Continual lifelong learning with neural networks: A review.
\newblock {\em Neural Networks}, 113:54--71, 2019.

\bibitem{pfeiffer2020adapterfusion}
Jonas Pfeiffer, Aishwarya Kamath, Andreas R{\"u}ckl{\'e}, Kyunghyun Cho, and
  Iryna Gurevych.
\newblock Adapterfusion: Non-destructive task composition for transfer
  learning.
\newblock {\em arXiv preprint arXiv:2005.00247}, 2020.

\bibitem{pham2021dualnet}
Quang Pham, Chenghao Liu, and Steven Hoi.
\newblock Dualnet: Continual learning, fast and slow.
\newblock {\em NeurIPS}, 2021.

\bibitem{pham2020contextual}
Quang Pham, Chenghao Liu, Doyen Sahoo, et~al.
\newblock Contextual transformation networks for online continual learning.
\newblock In {\em ICLR}, 2020.

\bibitem{prabhu2020gdumb}
Ameya Prabhu, Philip~HS Torr, and Puneet~K Dokania.
\newblock Gdumb: A simple approach that questions our progress in continual
  learning.
\newblock In {\em ECCV}, 2020.

\bibitem{raffel2020exploring}
Colin Raffel, Noam Shazeer, Adam Roberts, Katherine Lee, Sharan Narang, Michael
  Matena, Yanqi Zhou, Wei Li, and Peter~J Liu.
\newblock Exploring the limits of transfer learning with a unified text-to-text
  transformer.
\newblock {\em JMLR}, 21:1--67, 2020.

\bibitem{rao2019continual}
Dushyant Rao, Francesco Visin, Andrei Rusu, Razvan Pascanu, Yee~Whye Teh, and
  Raia Hadsell.
\newblock Continual unsupervised representation learning.
\newblock {\em NeurIPS}, 32, 2019.

\bibitem{rebuffi2017icarl}
Sylvestre-Alvise Rebuffi, Alexander Kolesnikov, Georg Sperl, and Christoph~H
  Lampert.
\newblock icarl: Incremental classifier and representation learning.
\newblock In {\em CVPR}, pages 2001--2010, 2017.

\bibitem{rusu2016progressive}
Andrei~A Rusu, Neil~C Rabinowitz, Guillaume Desjardins, Hubert Soyer, James
  Kirkpatrick, Koray Kavukcuoglu, Razvan Pascanu, and Raia Hadsell.
\newblock Progressive neural networks.
\newblock {\em arXiv preprint arXiv:1606.04671}, 2016.

\bibitem{serra2018overcoming}
Joan Serra, Didac Suris, Marius Miron, and Alexandros Karatzoglou.
\newblock Overcoming catastrophic forgetting with hard attention to the task.
\newblock In {\em ICML}, pages 4548--4557. PMLR, 2018.

\bibitem{shanahan2021encoders}
Murray Shanahan, Christos Kaplanis, and Jovana Mitrovi{\'c}.
\newblock Encoders and ensembles for task-free continual learning.
\newblock {\em arXiv preprint arXiv:2105.13327}, 2021.

\bibitem{shin2020autoprompt}
Taylor Shin, Yasaman Razeghi, Robert~L Logan~IV, Eric Wallace, and Sameer
  Singh.
\newblock Autoprompt: Eliciting knowledge from language models with
  automatically generated prompts.
\newblock {\em arXiv preprint arXiv:2010.15980}, 2020.

\bibitem{shokri2015privacy}
Reza Shokri and Vitaly Shmatikov.
\newblock Privacy-preserving deep learning.
\newblock In {\em Proc SIGSAC conference on computer and communications
  security}, 2015.

\bibitem{van2019three}
Gido~M Van~de Ven and Andreas~S Tolias.
\newblock Three scenarios for continual learning.
\newblock {\em arXiv preprint arXiv:1904.07734}, 2019.

\bibitem{vaswani2017attention}
Ashish Vaswani, Noam Shazeer, Niki Parmar, Jakob Uszkoreit, Llion Jones,
  Aidan~N Gomez, Lukasz Kaiser, and Illia Polosukhin.
\newblock Attention is all you need.
\newblock {\em NeurIPS}, 2017.

\bibitem{veniat2020efficient}
Tom Veniat, Ludovic Denoyer, and Marc'Aurelio Ranzato.
\newblock Efficient continual learning with modular networks and task-driven
  priors.
\newblock {\em arXiv preprint arXiv:2012.12631}, 2020.

\bibitem{wang2020k}
Ruize Wang, Duyu Tang, Nan Duan, Zhongyu Wei, Xuanjing Huang, Guihong Cao,
  Daxin Jiang, Ming Zhou, et~al.
\newblock K-adapter: Infusing knowledge into pre-trained models with adapters.
\newblock {\em arXiv preprint arXiv:2002.01808}, 2020.

\bibitem{wang2020learn}
Zifeng Wang, Tong Jian, Kaushik Chowdhury, Yanzhi Wang, Jennifer Dy, and
  Stratis Ioannidis.
\newblock Learn-prune-share for lifelong learning.
\newblock In {\em ICDM}, 2020.

\bibitem{wortsman2020supermasks}
Mitchell Wortsman, Vivek Ramanujan, Rosanne Liu, Aniruddha Kembhavi, Mohammad
  Rastegari, Jason Yosinski, and Ali Farhadi.
\newblock Supermasks in superposition.
\newblock {\em NeurIPS}, 33:15173--15184, 2020.

\bibitem{wu2019large}
Yue Wu, Yinpeng Chen, Lijuan Wang, Yuancheng Ye, Zicheng Liu, Yandong Guo, and
  Yun Fu.
\newblock Large scale incremental learning.
\newblock In {\em CVPR}, pages 374--382, 2019.

\bibitem{xiao2017fashion}
Han Xiao, Kashif Rasul, and Roland Vollgraf.
\newblock Fashion-mnist: a novel image dataset for benchmarking machine
  learning algorithms.
\newblock {\em arXiv preprint arXiv:1708.07747}, 2017.

\bibitem{yan2021dynamically}
Shipeng Yan, Jiangwei Xie, and Xuming He.
\newblock Der: Dynamically expandable representation for class incremental
  learning.
\newblock In {\em CVPR}, pages 3014--3023, 2021.

\bibitem{yoon2017lifelong}
Jaehong Yoon, Eunho Yang, Jeongtae Lee, and Sung~Ju Hwang.
\newblock Lifelong learning with dynamically expandable networks.
\newblock {\em arXiv preprint arXiv:1708.01547}, 2017.

\bibitem{zenke2017continual}
Friedemann Zenke, Ben Poole, and Surya Ganguli.
\newblock Continual learning through synaptic intelligence.
\newblock In {\em ICML}, 2017.

\bibitem{zeno2018task}
Chen Zeno, Itay Golan, Elad Hoffer, and Daniel Soudry.
\newblock Task agnostic continual learning using online variational bayes.
\newblock {\em arXiv preprint arXiv:1803.10123}, 2018.

\bibitem{zhang2021aggregating}
Zizhao Zhang, Han Zhang, Long Zhao, Ting Chen, , Sercan~Ö. Arık, and Tomas
  Pfister.
\newblock Nested hierarchical transformer: Towards accurate, data-efficient and
  interpretable visual understanding.
\newblock In {\em AAAI}, 2022.

\bibitem{zhao2022deep}
Tingting Zhao, Zifeng Wang, Aria Masoomi, and Jennifer Dy.
\newblock Deep bayesian unsupervised lifelong learning.
\newblock {\em Neural Networks}, 2022.

\end{thebibliography}
\end{document}